\documentclass[10pt,twocolumn,letterpaper]{article}

\usepackage[pagenumbers]{iccv} 

\usepackage{multirow} 
\usepackage{caption,subcaption,booktabs}
\usepackage{longtable}

\newcommand{\ra}[1]{\renewcommand{\arraystretch}{#1}} 
\usepackage{pifont}
\usepackage{makecell}

\newcommand{\bs}[1]{\boldsymbol{#1}}
\newcommand{\mc}[1]{\mathcal{#1}}
\newcommand{\mb}[1]{\mathbb{#1}}
\newcommand{\hbs}[1]{\hat{\boldsymbol{#1}}}
\usepackage[accsupp]{axessibility}

\definecolor{iccvblue}{rgb}{0.21,0.49,0.74}
\usepackage[pagebackref,breaklinks,colorlinks,allcolors=iccvblue]{hyperref}

\def\paperID{6949} 
\def\confName{ICCV}
\def\confYear{2025}

\title{UniEgoMotion: A Unified Model for Egocentric Motion Reconstruction, Forecasting, and Generation}

\author{
\vspace{4pt}  
\begin{tabular}{ccccc}
Chaitanya Patel$^{1}$ & & Hiroki Nakamura$^{2}$ & & Yuta Kyuragi$^{1,3}$ \\
Kazuki Kozuka$^{2}$ & & Juan Carlos Niebles$^{1}$ & & Ehsan Adeli$^{1}$
\end{tabular}\\
{\normalsize
$^{1}$Stanford University \quad
$^{2}$Panasonic Holdings Corporation \quad
$^{3}$Panasonic R\&D Company of America
} \\
{\small\url{https://chaitanya100100.github.io/UniEgoMotion/}}
}

\begin{document}



\twocolumn[{%
\renewcommand\twocolumn[1][]{#1}%
\maketitle
\begin{center}
    \centering
    \vspace{-3mm}
    \captionsetup{type=figure}
    \includegraphics[width=\textwidth]{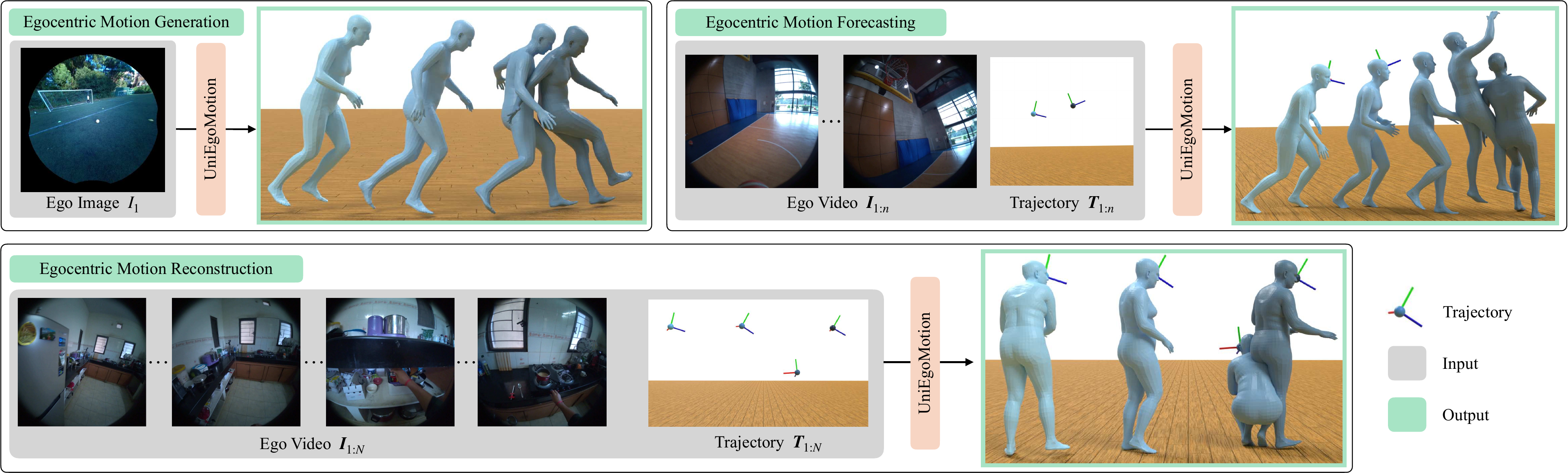}
    \captionof{figure}{
UniEgoMotion is a unified, scene-aware motion model designed for egocentric settings:
(1) It generates plausible future motion from a single egocentric image -- for example, predicting how you might take your shot on goal.
(2) It forecasts upcoming motion using past egocentric video and ego-device trajectory, showing how you could complete your run-up to score.
(3) It reconstructs accurate 3D motion from past egocentric observations, showing how you squatted down to reach the lower cabinet.
    }
    \label{fig:teaser}
\end{center}%
}]

\makeatletter
\makeatother
\maketitle

\begin{abstract}
Egocentric human motion generation and forecasting with scene-context is crucial for enhancing AR/VR experiences, improving human-robot interaction, advancing assistive technologies, and enabling adaptive healthcare solutions by accurately predicting and simulating movement from a first-person perspective.
However, existing methods primarily focus on third-person motion synthesis with structured 3D scene contexts, limiting their effectiveness in real-world egocentric settings where limited field of view, frequent occlusions, and dynamic cameras hinder scene perception.
To bridge this gap, we introduce Egocentric Motion Generation and Egocentric Motion Forecasting, two novel tasks that utilize first-person images for scene-aware motion synthesis without relying on explicit 3D scene. We propose UniEgoMotion, a unified conditional motion diffusion model with a novel head-centric motion representation tailored for egocentric devices.
UniEgoMotion’s simple yet effective design supports egocentric motion reconstruction, forecasting, and generation from first-person visual inputs in a unified framework.
Unlike previous works that overlook scene semantics, our model effectively extracts image-based scene context to infer plausible 3D motion.
To facilitate training, we introduce EE4D-Motion, a large-scale dataset derived from EgoExo4D, augmented with pseudo-ground-truth 3D motion annotations.
UniEgoMotion achieves state-of-the-art performance in egocentric motion reconstruction and is the first to generate motion from a single egocentric image.
Extensive evaluations demonstrate the effectiveness of our unified framework, setting a new benchmark for egocentric motion modeling and unlocking new possibilities for egocentric applications.
\end{abstract}

\section{Introduction}

Egocentric human motion reconstruction, forecasting, and generation are fundamental for AR/VR, assistive technologies, and healthcare applications.
The egocentric camera provides a personalized first-person perspective, enabling interactive and adaptive experiences.
Imagine you are learning to play soccer while wearing smart glasses. 
Egocentric motion \textbf{reconstruction} can help analyze and refine your kicking technique.
Egocentric motion \textbf{forecasting} (prediction) can show how you can continue your current run-up to execute a strike.
Egocentric motion \textbf{generation} can simulate how you might score a goal from your current position and angle.
These capabilities also extend to movement analysis in healthcare, aiding in gait assessment, early detection of neurological and vestibular disorders, and fall prediction.

Despite its potential, egocentric motion understanding remains a challenging problem.
The front-facing egocentric camera provides only partial visibility of the user’s body, forcing models to infer motion from a dynamic, first-person viewpoint with frequent occlusions and motion blur.
As a result, research on egocentric motion forecasting has been limited~\cite{yuan2019egopose}, and to our knowledge, no prior studies have explored egocentric motion generation.
Most context-aware motion forecasting and generation works assume explicit 3D scene context in the form of point cloud~\cite{mao2022contact, Scofano2023StagedCG, zheng2022gimo}, mesh~\cite{wang2020motionin3dscene}, voxel grid~\cite{hassan_samp_2021,cen2024text_scene_motion_llm}, signed distance field~\cite{xing2024eccv, PLACE:3DV:2020}, or object geometry~\cite{kulkarni2023nifty, li2024chois, corona2020context}.
Such 3D scene context is not available in many real-world egocentric applications.
While RGB videos are the most accessible modality to acquire scene context, very few works use image-based context for motion synthesis.
The most relevant prior works use a third-person scene-wide RGB image to forecast~\cite{cao2020scenecontext} or generate~\cite{wang2021sceneaware} human motion \textit{within} that scene.
However, these methods cannot generalize to egocentric settings where the wider scene context is unavailable due to the limited field of view and future motion may extend beyond the visible region, requiring strong motion priors.

To this end, we establish two novel tasks of scene-aware egocentric motion generation and forecasting only from egocentric images without requiring 3D scene context.
We introduce UniEgoMotion, a unified model for egocentric motion reconstruction, forecasting, and generation (see Figure \ref{fig:teaser}).
In particular, it can (1) reconstruct motion using an input ego video and ego-device's inertial SLAM trajectory, (2) forecast motion based on past egocentric inputs, and (3) generate motion from a single egocentric image.
Unlike prior works~\cite{li2023egoego,jiang2022avatarposer,yi2024egoallo,yuan2019egopose} that discard scene semantics, UniEgoMotion leverages egocentric image-based scene context to predict plausible and accurate 3D motion from an egocentric viewpoint.
To train UniEgoMotion, we present EE4D-Motion, a new dataset derived from the large-scale EgoExo4D~\cite{grauman2024egoexo4d} dataset. We augment EgoExo4D videos with paired pseudo-ground-truth 3D motion annotations using a comprehensive motion fitting pipeline, enabling training on in-context egocentric video-motion pairs.

At its core, UniEgoMotion is a transformer-based~\cite{vaswani2017attention} conditional motion diffusion model that enables flexible conditioning via cross-attention.
To establish scene context, it leverages a robust image encoder, initialized with strong pretraining~\cite{oquab2023dinov2}, to extract fine-grained visual features, enabling precise mapping of the visible environment while constructing a comprehensive prior of the unseen areas for holistic motion synthesis.
During training, we strategically mask conditioning inputs (ego images and 3D device trajectory) such that they support both egocentric reconstruction and generation during inference.
Egocentric forecasting is achieved through diffusion inpainting~\cite{lugmayr2022repaint} during inference, which utilizes the learned egocentric motion diffusion prior to predict future motion based on past motion reconstruction. 
Unlike pelvis-centric motion representations used in motion synthesis literature~\cite{guo2022generating,tevet2022mdm}, UniEgoMotion adopts a head-centric representation, making it more aligned with egocentric devices.
In addition to generation and forecasting, UniEgoMotion also outperforms state-of-the-art methods on egocentric motion reconstruction~\cite{li2023egoego,yi2024egoallo,jiang2022avatarposer} task.

In summary, we make the following contributions:
\begin{enumerate}
\item We introduce two novel tasks—Egocentric Motion Generation and Egocentric Motion Forecasting— expanding the scope of motion modeling for applications of wearable egocentric devices.
\item We propose UniEgoMotion, a novel unified egocentric motion model that performs reconstruction, forecasting, and generation in a single framework. It surpasses state-of-the-art baselines on egocentric motion reconstruction and, to our knowledge, is the first model to generate motion from a single egocentric image.
\item We present EE4D-Motion, a large-scale dataset of egocentric video-motion pairs, enabling the video-based scene context-aware human motion modeling.
\end{enumerate}

\section{Related Work}

\noindent\textbf{Scene-aware Motion Generation:}
Motion generation has been extensively studied in various settings, including character control \cite{holden2016deep,holden2017phase,ling2020character,henter2020moglow}, animation \cite{li2022ganimator,harvey2020robust,he2022nemf,starke2020local,raab2023modi}, action-to-motion synthesis \cite{guo2020action2motion,petrovich2021action,tevet2022mdm,wang2020learningact}, and more recently, text-to-motion generation \cite{guo2022generating,tevet2022mdm,petrovich2022temos,zhang2024motiondiffuse,ghosh2021synthesis,guo2022tm2t,zhang2023generatingfromtext,dabral2023mofusion,athanasiou2023sinc}. Scene-aware motion generation focuses on generating motion grounded in a given scene context. Most approaches condition motion generation on explicit 3D scene representations, such as scene point clouds \cite{zheng2022gimo,araujo2023circle,wang2022humanise}, meshes \cite{wang2020motionin3dscene}, voxel grids \cite{hassan_samp_2021,cen2024text_scene_motion_llm,starke2019neuralscene,jiang2024scaling}, signed distance fields \cite{PLACE:3DV:2020,hassan2019prox,yi2024tesmo,wang2024move}, or specific object geometries \cite{kulkarni2023nifty,li2024chois,corona2020context,zhang2022couch}. Some works leverage 3D scene context for navigational motion generation \cite{zhang2022wanderings,wang2020motionin3dscene}, while others model human-scene \cite{araujo2023circle,wang2022humanise,cen2024text_scene_motion_llm,hassan_samp_2021,jiang2024scaling,starke2019neuralscene,PLACE:3DV:2020,hassan2019prox,yi2024tesmo,wang2024move} and human-object interactions \cite{kulkarni2023nifty,li2024chois,corona2020context,zhang2022couch,starke2019neuralscene}.

However, capturing high-quality 3D scene data requires complex setups or extensive offline reconstruction \cite{gu2024egolifter,tschernezki2023epicfields,pan2023adt}, making it impractical for real-world egocentric applications. In contrast, RGB images are easily accessible but pose challenges in extracting relevant scene context due to their limited field of view, dynamic motion, and occlusions. Yet, motion generation from image-based scene context remains underexplored, with \cite{wang2021sceneaware} being the only work proposing a two-stage GAN-based \cite{goodfellow2014gan} model to generate human motion from a single wide-scene RGB image. We tackle a more challenging yet practical egocentric setting by introducing the egocentric motion generation task, which generates context-aware motion from a single egocentric image.

\noindent \textbf{Motion Forecasting:}
Motion forecasting, i.e., predicting future motion based on past motion, has been widely explored in a context-independent setting. Approaches range from traditional  models~\cite{lehrmann2014efficient,wang2005gaussian,taylor2006modeling} to modern deep learning, including MLPs~\cite{guo2023back}, RNNs~\cite{chiu2019action,fragkiadaki2015recurrent,jain2016structural,martinez2017human,wang2019imitation}, graph convolutional networks~\cite{dang2021msr,guo2022multi}, Transformers~\cite{aksan2021spatio,cai2020learning,mao2020history,martinez2021posetransformer}, RL controllers~\cite{yuan2019egopose}, and diffusion models~\cite{barquero2023belfusion}.

Similar to motion generation, scene-aware motion forecasting assumes access to a clean 3D scene as context to predict future motion~\cite{corona2020context,Scofano2023StagedCG,zheng2022gimo,mao2022contact,xing2024eccv}, with a few exceptions~\cite{cao2020scenecontext,yuan2019egopose}. \cite{cao2020scenecontext} uses a single wide-scene RGB image along with past poses to predict future motion. It proposes a three-stage method: stochastic goal prediction via a VAE~\cite{kingma2013vae}, deterministic trajectory prediction, and deterministic pose generation. \cite{yuan2019egopose} applies an RL-based controller for motion forecasting from egocentric devices, though its evaluation is limited to simple actions like walking and running. In contrast, we address scene-aware motion forecasting in an egocentric setting using diffusion modeling~\cite{ho2020denoising}, enabling more diverse and complex motion predictions.

\noindent \textbf{Egocentric Motion Reconstruction:}
Unlike downward-facing camera setting~\cite{liu2023egohmr,park2023domain,rhodin2016egocap,tome2020selfpose,wang2022estimating}, egocentric motion reconstruction focuses on front-facing ego cameras, which are more common in publicly available devices~\cite{engel2023projectaria}. This introduces significant challenges due to limited body visibility and requires strong motion priors.
Many works use simulation-based physical motion priors~\cite{yuan2019egopose,luo2021kinpoly,yuan2018estimation} or diffusion motion priors~\cite{li2023egoego,yi2024egoallo,guzov2024hmd2,jiang2022avatarposer} learned from large motion datasets. However, \cite{yuan2019egopose,li2023egoego} use input video to compute optical flow to estimate the ego-device’s 3D trajectory, discarding valuable scene context. \cite{yi2024egoallo,castillo2023bodiffusion,jiang2022avatarposer} depend exclusively on accurate ego-device trajectory computed via SLAM, without incorporating scene context. These approaches are suboptimal for activities where head motion is minimal, such as cooking or playing a music instrument.
The closest work to ours is~\cite{guzov2024hmd2}, which integrates scene point clouds and ego-image features for scene-aware egocentric motion reconstruction. In contrast, our approach does not rely on point cloud input; instead we leverage only egocentric images to capture scene context.
Some concurrent works~\cite{EgoLM, wang2025ego4o} finetune language models on motion and text from the Nymeria~\cite{ma2024nymeria} dataset, conditioning on egocentric inputs, for autoregressive motion reconstruction and understanding.
Note that prior works often refer to this task as `motion generation' due to its generative nature. However, we differentiate between egocentric motion generation and reconstruction based on available sensory information.

\section{Method}

\begin{figure*}[h]
\centering
\includegraphics[width=0.9\textwidth]{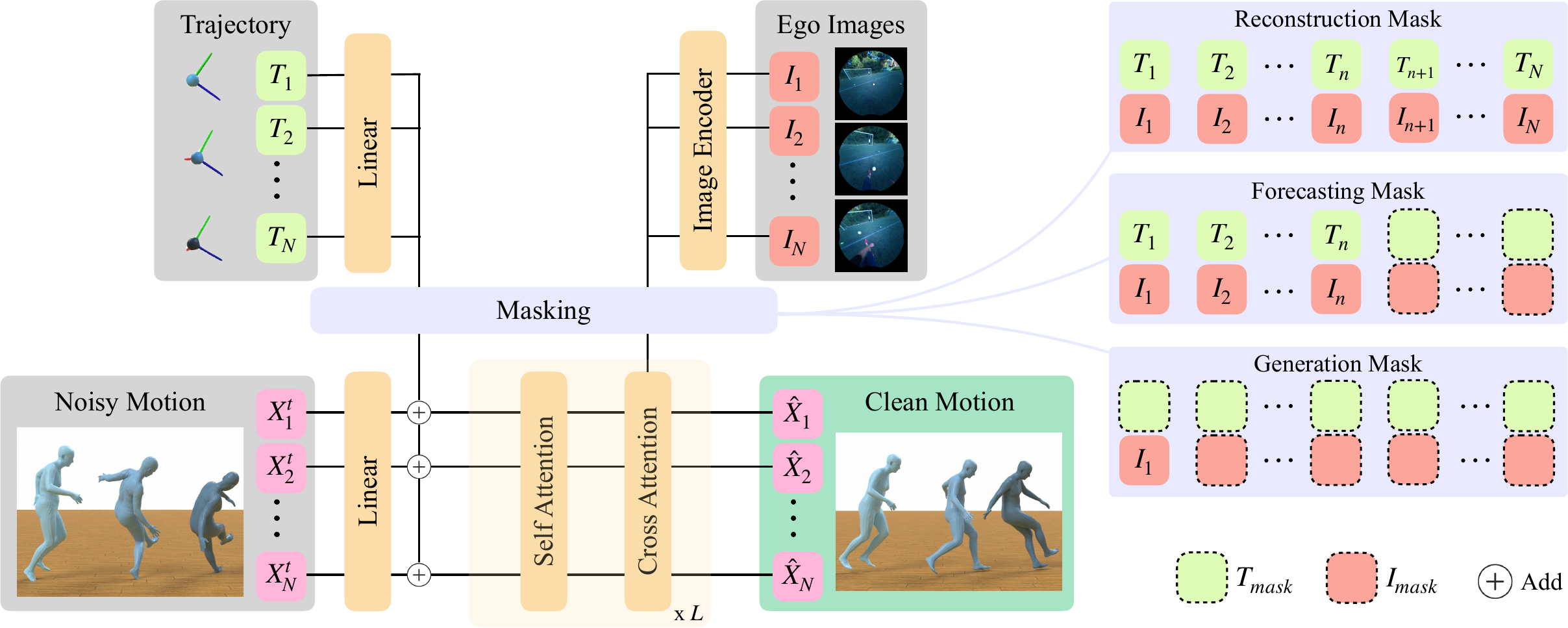}
\caption{
Overview of a denoising step in UniEgoMotion. The input noisy motion $\bs{X}^t_{1:N}$ is denoised using a transformer decoder network conditioned on the ego-device trajectory $\bs{T}_{1:N}$ and egocentric images $\bs{I}_{1:N}$. A robust image encoder is used to extract fine-grained scene context from the images. During training, conditioning inputs are randomly replaced with learnable mask tokens to simulate three tasks: egocentric reconstruction, forecasting, and generation. During inference, the learned mask tokens are used in place of any missing conditioning input, allowing a single model to perform all three tasks consistently.
}
\label{fig:overview}
\end{figure*}

\subsection{Problem Formulation}
\label{subsec:problem_formulation}

Let $\bs{I}_{1:N} = (I_1, I_2, \cdots, I_N)$ denote a sequence of egocentric video frames captured by a head-mounted camera, where each RGB frame $I_i \in \mb{R}^{H\times W\times 3}$.
Let $\bs{T}_{1:N} = (T_1, T_2, \cdots, T_N)$ represent the camera's 6-DOF trajectory.
Modern wearable devices equipped with state-of-the-art inertial SLAM systems~\cite{engel2023projectaria} can compute $\bs{T}_{1:N}$ in real-time.
Let $\bs{X}_{1:N} = (X_1, X_2, \cdots, X_N)$ denote the 3D human motion of the user wearing the camera.
Each pose $X_i$ at timestamp $i$ is defined by SMPL-X~\cite{pavlakos2019smplx} parameters $X_i = (R^r_i, t^r_i,  \theta_i, \beta_i)$ where $R^r_i \in \mb{R}^3$ and $t^r_i \in \mb{R}^3$ denotes root joint's global rotation and translation (at pelvis), $\theta_i \in \mb{R}^{21\times 3}$ denotes the local joint angles of the kinematic skeleton with 21 joints, and $\beta_i \in \mb{R}^{10}$ denotes body shape which remains constant over time ($\beta_i = \beta_j$ for all $i, j$). Using this notation, we define three egocentric motion tasks:

\textbf{Egocentric Motion Generation} aims to synthesize plausible future motion from a single egocentric image. Formally, this task involves sampling from $p(\bs{X}_{1:N} \vert I_1)$, where the front-facing egocentric image $I_1$ provides a personalized scene context. This problem is more challenging than 3D scene-aware motion generation because, without explicit 3D scene input, the model must infer geometric information from the visible scene and make plausible assumptions about occluded regions. For instance, if the right corner of a soccer field is visible, the model must reason about the likely position of the goal net. Additionally, the model must infer the ongoing action without explicit action labels or textual prompts. For example, if the egocentric image captures raised hands holding a basketball, the most plausible motion is taking a shot at the basket.

\textbf{Egocentric Motion Forecasting} predicts future motion given past egocentric observations, formulated as sampling from $p(\bs{X}_{n+1:N} \vert \bs{I}_{1:n}, \bs{T}_{1:n})$ where $n < N$.
This process implicitly involves reconstructing past motion, represented as $p(\bs{X}_{1:n} \vert \bs{I}_{1:n}, \bs{T}_{1:n})$, followed by traditional pose forecasting.
Thus, it can also be formulated as $p(\bs{X}_{n+1:N} \vert \bs{I}_{1:n}, \bs{X}_{1:n})$.
\footnote{It is reasonable to assume that the egocentric device is placed at a fixed location with respect to the user's head and device location $T_i$ can be derived from pose $X_i$ using a known transform.}
In this paper, we adopt the former definition.
This task is easier than generation as the model benefits from the past observations, providing additional context and constraints on plausible future motion.

\textbf{Egocentric Motion Reconstruction} aims to recover motion from an egocentric video and the corresponding ego camera trajectory. It is formulated as sampling from $p(\bs{X}_{1:N} \vert \bs{I}_{1:N}, \bs{T}_{1:N})$. Compared to generation and forecasting, reconstruction is relatively easier due to its strong frame-aligned conditioning. Prior works~\cite{li2023egoego,yi2024egoallo,jiang2022avatarposer,castillo2023bodiffusion,guzov2024hmd2} often refer to this as a `generation' task because of the body's partial visibility in egocentric views. However, we adopt the term `reconstruction' based on its frame-wise aligned conditioning to distinguish it from egocentric generation and forecasting.

\subsection{Diffusion Motion Modeling}
Most diffusion-based motion models follow~\cite{ho2020denoising, tevet2022mdm} and train a conditional diffusion model~\cite{ho2022cfg}
$$\hbs{X} = \mc{M}\left(\bs{X}^t, \; t, \; \bs{C}; \; \Theta\right)$$
where $\bs{X}^t$ is a noised version of the clean motion $\bs{X}$, $\hbs{X}$ is the predicted clean motion,  $t$ is the diffusion timestep, $\bs{C}$ represents optional conditioning inputs, and $\Theta$ are the learnable model parameters. For clarity, motion frame indices are omitted i.e. $\bs{X} = \bs{X}_{1:N}$.
Following~\cite{ho2020denoising}, $\bs{X}$ is sampled using a forward Gaussian diffusion process 
$$q_t(\bs{X}^t | \bs{X}) = \mc{N}(\bs{X}^t; \sqrt{\bar{\alpha}_t} \bs{X}, (1 - \bar{\alpha}_t)\bs{I})$$
where  $\bar{\alpha}_t$ defines a monotonically increasing noise schedule. The model $\mc{M}(\cdot \vert \cdot\;; \Theta)$ learns the reverse diffusion process by minimizing the following denoising loss.
$$ \mc{L} = \mb{E}_{ t \in [1, t_{max}] , \bs{X}^t \sim q_t(\cdot | \bs{X})} \left[ \left\| \bs{X} -     \mc{M}\left(\bs{X}^t, \; t, \; \bs{C} \right)     \right\|_2^2 \right] $$
The conditioning input $\bs{C}$ can include text prompts, action labels, 3D scene or object geometry, or other relevant features, depending on the task.
During inference, sampling starts from random Gaussian noise $\bs{X}^{t_{max}} \sim \mc{N}(\bs 0, \bs I)$, and iteratively denoises through
\begin{align*}
\bs{X}^{t-1} = \mc{M}(\bs{X}^t, t, \bs{C}) + \epsilon_t 
\label{eq:diff_sampling}
\end{align*}
until $t=1$, ultimately generating a clean sequence $\bs{X}^0$.

\subsection{UniEgoMotion}

Instead of training separate models for egocentric reconstruction, forecasting, and generation, we use a unified approach where the conditioning $\bs{C}$ is adapted to the specific task, as described in~\S\ref{subsec:problem_formulation}.
Recent egocentric motion reconstruction methods~\cite{li2023egoego,yi2024egoallo,castillo2023bodiffusion} fit into this framework by setting $\bs{C} = \bs{T}_{1:N}$.
However, we leverage the fact that conditional diffusion models trained with classifier-free guidance~\cite{ho2022cfg} support sampling from both conditional and unconditional distribution.
While~\cite{ho2022cfg} used unconditional generation to balance sample quality and diversity, we employ it specifically for egocentric motion generation.
During training, we randomly set $\bs{C} = \{\bs{T}_{1:N}, \bs{I}_{1:N}\}$ to simulate the reconstruction task and $\bs{C} = \{I_1\}$ to simulate the generation task, covering the both extremes.

Forecasting can be trained by setting $\bs{C} = \{\bs{T}_{1:n}, \bs{I}_{1:n}\}$ where $n < N$.
For the reconstruction-then-forecasting approach, diffusion repainting~\cite{lugmayr2022repaint} can also be applied during inference.
In particular, given inputs $\{\bs{T}_{1:n}, \bs{I}_{1:n}\}$, we first reconstruct the observed motion (using egocentric reconstruction) as $\bs{X}_{1:n} \sim \mc{M}(\cdot \vert \bs{T}_{1:n}, \bs{I}_{1:n})$. We then condition on $\bs{C} = \{\bs{T}_{1:n}, \bs{I}_{1:n}\}$ and sample the full motion sequence $\hbs{X}_{1:N}$, enforcing consistency by overwriting the known frames at each step as the following.
\begin{align*}
\hbs{X}_{1:N} \leftarrow{} \text{concat}(\bs{X}_{1:n}, \hbs{X}_{n+1:N}) 
\end{align*}

\subsection{Architecture}

We implement UniEgoMotion using a transformer-based~\cite{vaswani2017attention} architecture to denoise noisy motion input $\bs{X}^t_{1:N}$. Each motion input $X_i$ is projected into a latent vector via a linear layer, $f_X(X_i)$, and then processed by multiple transformer decoder layers.
For conditioning, we use full conditioning $\bs{C}=\{ \bs{T}_{1:N}, \bs{I}_{1:N} \}$ during reconstruction.
In generation mode, where $\bs{C} = \{I_1\}$, we use a learnable mask inputs to create full conditioning.
In particular, we set $T_i = T^\text{mask}$ for all $i \in \{1, \cdots, N\}$ and $I_i = I^\text{mask}$ for all $i \in \{2, \cdots, N\}$.
Forecasting conditioning is processed in a similar manner.
Each $T_i$ and $I_i$ is projected into a latent vector using $f_T(T_i)$ and $f_I(I_i)$, where $f_T$ is a linear layer and $f_I$ is a ViT-based image encoder.
$f_T(T_i)$ is added to $f_X(X_i)$ before passing through the transformer, while $f_I(I_i)$ is incorporated via a cross-attention mechanism.

Unlike prior works~\cite{li2023egoego,yi2024egoallo} that discard semantic information from ego images, we integrate fine-grained scene-aware features through $f_I$.
We show that the choice of $f_I$ significantly impacts the accuracy and fidelity of motion prediction. Training $f_I$ from scratch is suboptimal, as extracting scene context from images is a challenging problem in itself.
To address this, we leverage a pretrained DINOv2~\cite{oquab2023dinov2} to initialize $f_I$, training only the projector network. Our results show that its fine-grained features from~\cite{oquab2023dinov2} yield significant improvements over other strong image encoders~\cite{radford2021clip, pei2024egovideo}.

\subsection{Motion Representation}

Although SMPL-X parameters $X_i = (R^r_i, t^r_i,  \theta_i, \beta_i)$ are sufficient to represent 3D body motion, they are not ideal for learning~\cite{guo2022generating,yi2024egoallo}. The global root trajectory $(R^r_i, t^r_i)$, defined at the pelvis, fails to exploit motion redundancies, requiring the model to learn all directions explicitly. Moreover, a mismatch exists between the egocentric conditioning inputs $(T_i, I_i)$ and the pelvis-centric SMPL-X parameters, complicating motion reasoning. Using local joint angles further forces to learn complex forward kinematics, often leading to artifacts like foot-floor penetration and sliding.

To address these issues, we adopt a head-centric representation. We transform $X_i$ into $(M^h_i, \bs{M}^j_i)$ using forward kinematics where  $M^h_i \in \mb{R}^{4\times 4}$ is the global SE(3) transform of the head joint, and $\bs{M}^j_i \in \mb{R}^{21 \times 4\times 4}$ are those of other joints. This removes joint dependencies in the kinematic chain. Next, we derive a canonical reference frame ${}_cM_i$ per frame by eliminating pitch, roll, and height relative to the floor, ensuring that ${}_cM_i$ captures the head’s global trajectory projected onto the floor. Motion $(M^h_i, \bs{M}^j_i)$ is then expressed as $({}_cM_i, {}_cM_i \odot M^h_i, {}_cM_i \odot \bs{M}^j_i)$ where the latter terms encode local canonicalized pose information. For trajectory invariance, we represent ${}_cM_i$ as its residual relative to the previous frame. For more details, please refer to the supplementary material.
%
%
While~\cite{yi2024egoallo} adopts a similar canonicalization scheme, it preserves the kinematic chain, resulting in severe foot-floor penetration and floating artifacts. Our experiments validate the effectiveness of our motion representation against ~\cite{yi2024egoallo}.

\begin{figure*}[h]
\centering
\vspace{-2mm}
\includegraphics[width=\textwidth]{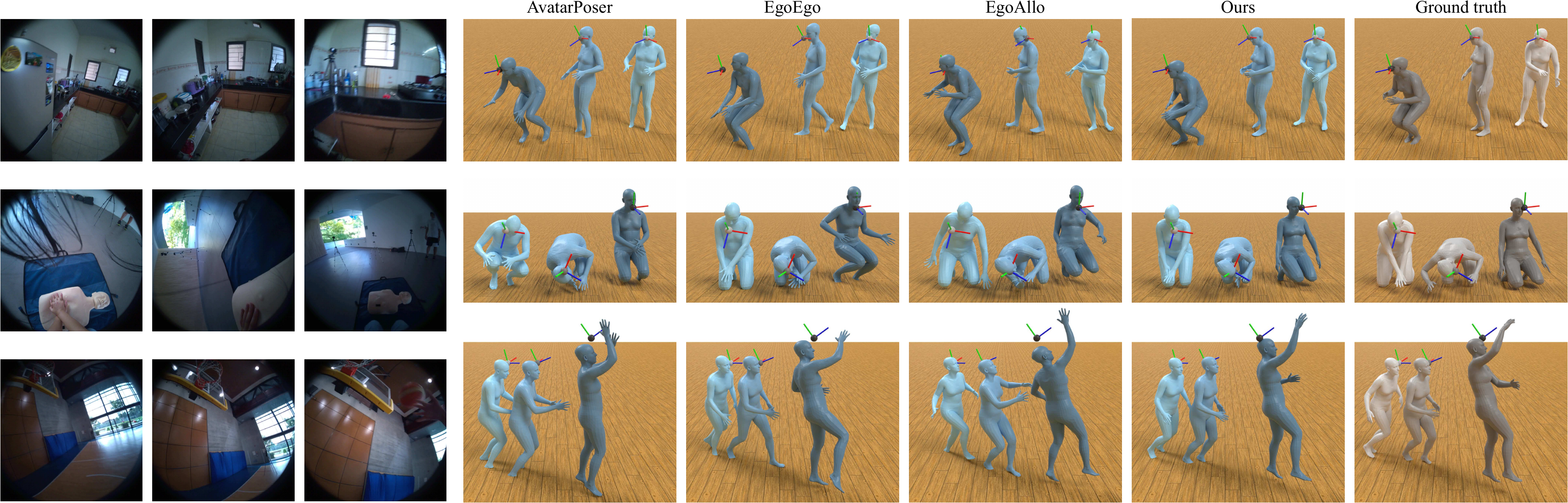}
\caption{Qualitative comparison of Egocentric Reconstruction. The input egocentric images are shown on the left, with the corresponding ego-device trajectory visualized alongside the predictions. Baseline methods exhibit floating motion, floor penetration, and inaccurate joint localization, whereas UniEgoMotion generates reconstructions that closely align with the ground truth.}
\vspace{-2mm}
\label{fig:recon}
\end{figure*}

\begin{table*}\centering \footnotesize
\setlength{\tabcolsep}{7.0pt}
\ra{1.2} 
\caption{\textbf{Egocentric Motion Reconstruction}: Comparison  of the reconstruction capabilities of UniEgoMotion with prior works (top).  Ablation on UniEgoMotion's model design for the reconstruction task (bottom). Note that the vanilla UniEgoMotion model uses transformer decoder architecture, head-centric motion representation, and DINOv2 visual encoder.}
\vspace{-3mm}
\begin{tabular}{lccccccccc}
\toprule
\thead{Method} & \thead{Head Rot.\\ Err.} & \thead{Head Trans.\\Err.} &  \thead{MPJPE} & \thead{MPJPE-PA} & \thead{MPJPE-H} & \thead{Foot\\Slide} & \thead{Foot\\Contact} & \thead{Semantic\\Sim.($\uparrow$)} & \thead{FID} \\
\cmidrule{1-10}
AvatarPoser~\cite{jiang2022avatarposer} & - & - & $0.116$ & $0.068$ & $0.240$ &	$7.85$ & $0.042$ &	$0.872$ & $0.082$ \\
EgoEgo~\cite{li2023egoego} & - & - & $0.130$ &	$0.075$ &	$0.272$ &	$3.90$ &	$0.033$ &	$0.858$ &	$0.068$ \\
EgoAllo~\cite{yi2024egoallo} & - &	- &	$0.163$ &	$0.071$ &	$0.273$ &	$4.10$ &	$0.056$ &	$0.885$ &	$0.043$ \\ 
UniEgoMotion & $\mathbf{0.260}$ &	$0.058$ &	$\mathbf{0.100}$ &	$\mathbf{0.053}$ &	$\mathbf{0.180}$ &	$\mathbf{3.62}$ &	$0.027$ &	$\mathbf{0.918}$ &	$0.027$ \\
\cmidrule{1-10}
Transformer Encoder & $0.280$ &	$0.076$ &	$0.115$ &	$0.056$ &	$0.189$ &	$4.48$ &	$0.029$ &	$0.912$ &	$\mathbf{0.017}$ \\
1D U-Net & $0.338$ &	$0.109$ &	$0.145$ &	$0.061$ &	$0.224$ &	$5.86$ &	$0.032$ &	$0.900$ &	$0.019$ \\
Global Motion Repre. & $0.275$ &	$\mathbf{0.051}$ &	$0.101$ &	$0.057$ &	$0.192$ &	$3.73$ &	$\mathbf{0.025}$ &	$0.912$ &	$0.024$ \\
Pelvis-centric Repre. & $0.398$ &	$0.138$ &	$0.166$ &	$0.054$ &	$0.241$ &	$3.66$ &	$0.028$ &	$0.909$ &	$0.030$  \\
CLIP encoder & $0.269$ &	$0.062$ &	$0.107$ &	$0.056$ &	$0.191$ &	$4.03$ &	$0.032$ &	$0.911$ &	$0.021$  \\
EgoVideo encoder & $0.332$ &	$0.101$ &	$0.132$ &	$0.060$ &	$0.211$ &	$4.73$ &	$0.032$ &	$0.897$ &	$0.041$ \\
\bottomrule
\end{tabular}
\label{tab:recon_compare}
\vspace{-4mm}
\end{table*}

\subsection{EE4D-Motion Dataset}
To train UniEgoMotion, we process EgoExo4D dataset~\cite{grauman2024egoexo4d} and develop EE4D-Motion dataset that provides synchronized egocentric videos and pseudo-ground-truth 3D motion data. Since existing datasets either lack paired egocentric videos or motion annotations, we develop a processing pipeline to fit SMPL-X~\cite{bogo2016smplify} to EgoExo4D sequences. Our approach refines initial pose estimates through multi-view optimization, sequence-level smoothing, and quality filtering, producing 110+ hours of 3D-accurate motion data for real-world activities. Please refer to suppl. material for more details on EE4D-Motion.


\begin{figure*}[h]
\centering
\vspace{-2mm}
\includegraphics[width=0.9\textwidth]{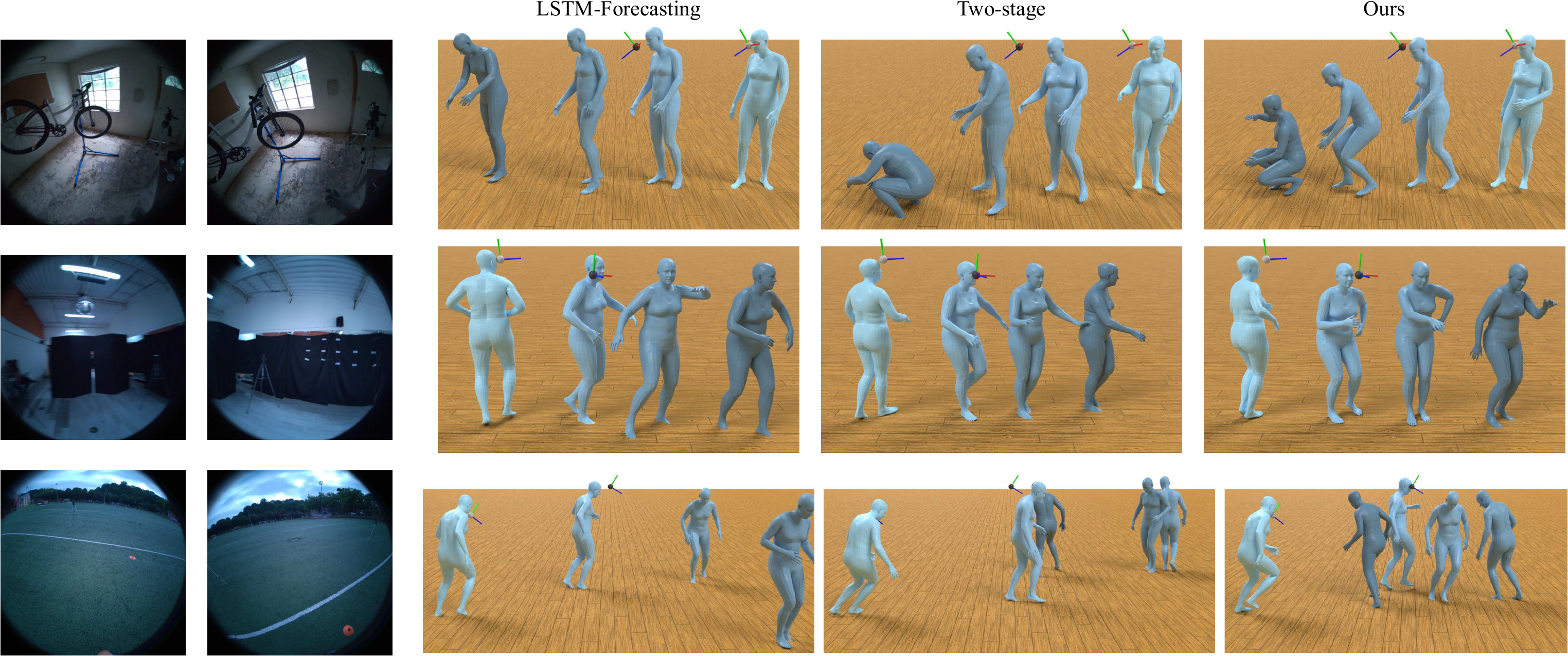}
\vspace{-2mm}
\caption{
Qualitative comparison of Egocentric Forecasting for predicting future motion using the first 2 seconds of egocentric video and trajectory input. The LSTM baseline predicts an average future motion and suffers from foot sliding, while the Two-stage baseline produces damped motion. In contrast, our model successfully predicts complex motions, such as squatting down to repair a bike tire (top), performing a salsa dance (middle), and executing a dribbling drill around a dome cone (bottom).
}
\label{fig:fore}
\end{figure*}

\begin{table*}\centering \footnotesize
\setlength{\tabcolsep}{4.0pt}
\ra{1.2} 
\caption{
The baselines and ablations are evaluated on egocentric motion forecasting (left) and generation (right). The metrics reported include \textbf{J}: MPJPE, \textbf{J-PA}: MPJPE-PA, \textbf{J-H}: MPJPE-H, \textbf{FS}: Foot Slide, \textbf{FC}: Foot Contact, and \textbf{SS}: Semantic Similarity. MPJPE metrics are computed over the first two seconds of future predictions (0-2s for generation and 2-4s for forecasting). $^*$Two-stage baseline replicates the trajectory-to-motion prediction framework used in prior works on image-based motion forecasting~\cite{cao2020scenecontext} and motion generation~\cite{wang2021sceneaware}.
}
\vspace{-3mm}
\begin{tabular}{l ccccccc|ccccccc}
\toprule
\multicolumn{1}{c@{}}{} & \multicolumn{7}{c@{}}{Egocentric Motion Forecasting} & \multicolumn{7}{c@{}}{Egocentric Motion Generation} \\
\cmidrule{2-15}
\thead{Method} & \thead{J\\(2-4s)} & \thead{J-PA\\(2-4s)} & \thead{J-H\\(2-4s)} & \thead{FS} & \thead{FC} & \thead{SS ($\uparrow$)} & \thead{FID} &
\thead{J\\(0-2s)} & \thead{J-PA\\(0-2s)} & \thead{J-H\\(0-2s)} & \thead{FS} & \thead{FC} & \thead{SS ($\uparrow$)} & \thead{FID} \\
\cmidrule{1-15}
LSTM & $0.238$ &	$\mathbf{0.066}$ &	$0.330$ &	$7.23$ &	$0.031$ &	${0.849}$ &	$0.058$  
& $\mathbf{0.216}$ &	$\mathbf{0.067}$ &	$\mathbf{0.308}$ &	$6.83$ &	$0.028$ &	$0.809$ &	$0.090$ \\
Two-stage$^*$~\cite{wang2021sceneaware,cao2020scenecontext} & $0.253$ &	$0.072$ &	$0.361$ &	$3.55$ &	$\mathbf{0.026}$ &	$\mathbf{0.850}$ &	$\mathbf{0.038}$ 
& $0.222$ &	$0.072$ &	$0.323$ &	$4.35$ &	$0.026$ &	$\mathbf{0.822}$ &	$0.037$ \\
UniEgoMotion & $\mathbf{0.206}$ &	$0.071$ &	$\mathbf{0.308}$ &	$\mathbf{2.60}$ &	$\mathbf{0.026}$ &	${0.849}$ &	$0.047$ 
& $0.226$ &	$0.070$ &	$0.321$ &	$\mathbf{2.89}$ &	$\mathbf{0.025}$ &	${0.817}$ &	$0.043$  \\
\cmidrule{1-15}
Transformer Encoder & $0.213$ &	$0.073$ &	$0.315$ &	$3.13$ &	$0.027$ &	$0.846$ &	${0.041}$ 
& $0.231$ &	$0.072$ &	$0.330$ &	$3.40$ &	$0.026$ &	$0.814$ &	$\mathbf{0.034}$ \\
1D U-Net &  $0.239$ &	$0.075$ &	$0.346$ &	$4.11$ &	$0.027$ &	$0.840$ &	$0.056$
& $0.259$ &	$0.079$ &	$0.360$ &	$4.06$ &	$0.029$ &	$0.802$ &	$0.035$ \\
Global Motion Repre. & $0.293$ &	$0.076$ &	$0.405$ &	$3.16$ &	$0.027$ &	$0.841$ &	$0.046$ 
& $0.228$ &	$0.072$ &	$0.328$ &	$3.65$ &	$0.025$ &	$0.821$ &	$0.035$ \\
Pelvis-centric Repre. & $0.245$ &	$0.074$ &	$0.354$ &	$3.56$ &	$0.030$ &	$0.838$ &	$0.042$  
& $0.232$ &	$0.071$ &	$0.327$ &	$3.94$ &	$0.029$ &	$0.814$ &	$0.039$ \\
CLIP encoder & $0.214$ &	$0.073$ &	$0.315$ &	$3.75$ &	$0.030$ &	$0.844$ &	$0.043$ 
& $0.238$ &	$0.073$ &	$0.333$ &	$3.57$ &	$0.028$ &	$0.816$ &	$0.037$ \\
EgoVideo encoder & $0.228$ &	$0.077$ &	$0.331$ &	$3.65$ &	$0.030$ &	$0.835$ &	$0.060$ 
& $0.236$ &	$0.074$ &	$0.322$ &	$3.50$ &	$0.030$ &	$0.812$ &	$0.059$ \\
\bottomrule
\end{tabular}
\label{tab:fore_gen_compare}
\vspace{-3mm}
\end{table*}

\begin{figure*}[h]
\centering
\vspace{-2mm}
\includegraphics[width=0.9\textwidth]{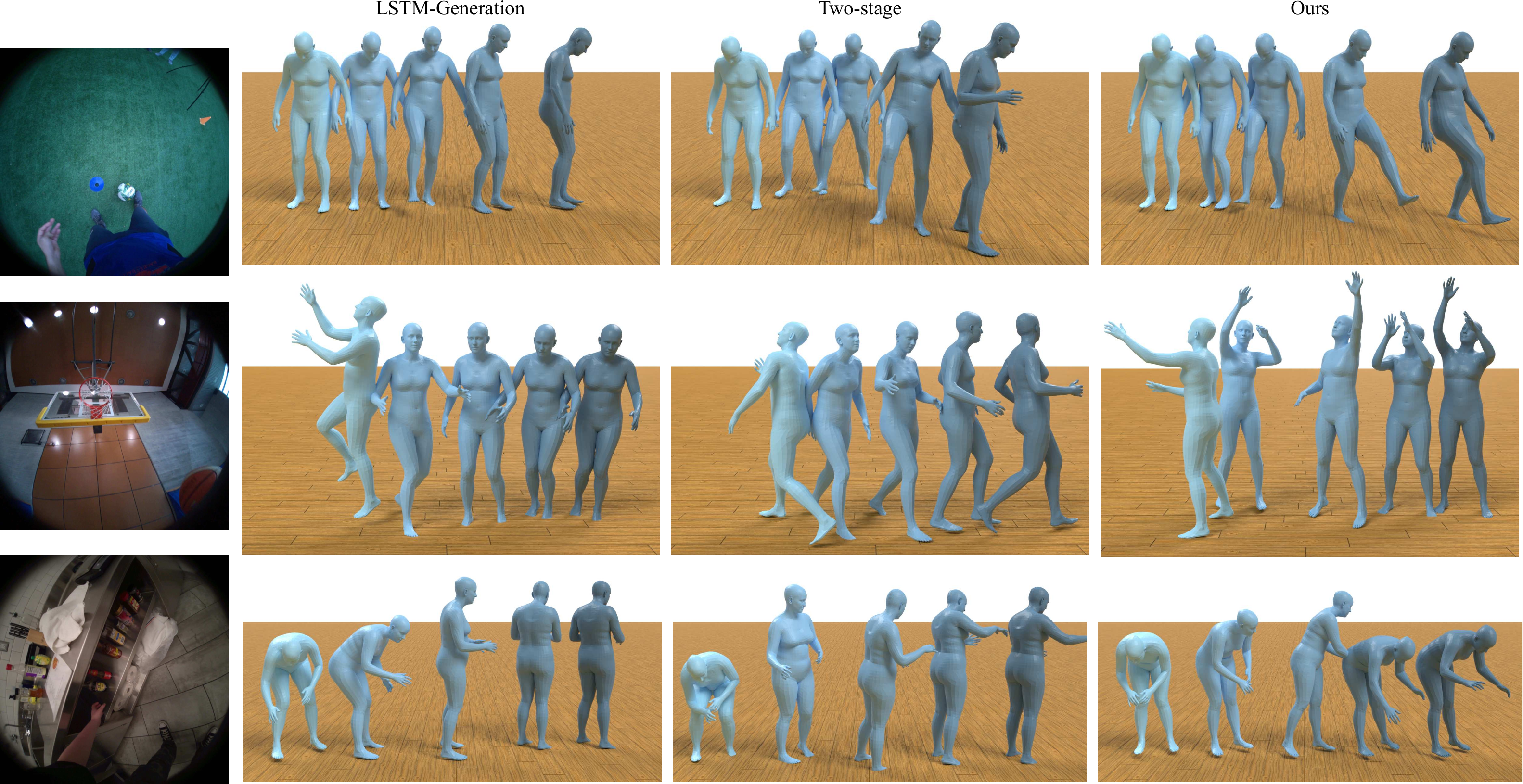}
\vspace{-2mm}
\caption{
Qualitative comparison of Egocentric Motion Generation from a single egocentric image input.
Compared to the LSTM and Two-stage baseline, our model leverages the fine-grained image features for more accurate motion generation, demonstrating soccer juggling (top), a basketball shooting drill (middle), and interaction with the lower cabinet on the \textit{left} side of the person.
}
\vspace{-2mm}
\label{fig:gen}
\end{figure*}

\section{Experiments}

We follow the official split of the EgoExo4D dataset~\cite{grauman2024egoexo4d} to partition the EE4D-Motion dataset into training and validation sets based on capture takes. UniEgoMotion and other baselines are trained on 8-second video clips sampled at 10fps ($N=80$), with clips extracted every 2 seconds, resulting in a total of 143K training samples. For evaluation, we sample similar video clips every 20 seconds, yielding 4400 validation samples. For forecasting, we predict 6 seconds into the future after observing first 2 seconds ($n=20$) of egocentric inputs. We train a single UniEgoMotion model and evaluate it across all three tasks.

\subsection{Metrics}

We employ several metrics to evaluate motions in both 3D and semantic space.
\textbf{MPJPE} calculates the mean per-joint positional error (in meters) of 22 body joints.
\textbf{MPJPE-PA} applies Procrustes analysis to align ground truth and predicted motions per frame before computing MPJPE, measuring the accuracy of local pose predictions.
\textbf{MPJPE-H} calculates the mean per-joint positional error (in meters) of hand joints.
\textbf{Head Rotation Error} and \textbf{Head Translation Error} measure the rotation error and translation error (in meters) of the head joint, respectively, capturing the model’s ability to adhere to head-aligned conditioning inputs for the reconstruction task. Rotation error is calculated as the frobenius norm of the difference rotation matrix~\cite{li2023egoego}.
\textbf{Foot Sliding}~\cite{he2022nemf} quantifies the extent of foot sliding when the foot is close to the ground.
\textbf{Foot Contact} computes the average separation (in meters) between foot and ground. It quantifies both floating and floor penetration.
\textbf{Semantic Similarity} evaluates the similarity between generated and ground-truth motions similar to CLIP-score~\cite{hessel2021clipscore}. Specifically, we leverage the motion encoder from TMR~\cite{guo2022tm2t} to embed motion into a latent space and compute the cosine distance between embeddings.
\textbf{FID} measures the distributional discrepancy between generated and ground-truth motions in the latent space, akin to vanilla FID for images.

For generation and forecasting, we compute the MPJPE and MPJPE-PA metrics only for the first 2 seconds of prediction, as beyond that, generated motions may remain valid despite exhibiting large joint errors.
FID and Semantic Similarity compares motions in a semantic latent space and offers a more meaningful evaluation of the motion quality and scene-relevance respectively.
Foot Slide and Foot Contact capture physical realism of the predicted motion.

\subsection{Baselines}

We compare our egocentric motion reconstruction with task-specific prior works: EgoEgo~\cite{li2023egoego}, EgoAllo~\cite{yi2024egoallo}, and AvatarPoser~\cite{jiang2022avatarposer}. To ensure a fair evaluation, we retrain each method on the EE4D-Motion dataset using publicly available code. None of these baselines use semantic information from egocentric images for prediction. We compare these baselines on the reconstruction task and further ablate their design choices within our UniEgoMotion framework in a consistent manner.
Since there are no direct baselines for egocentric motion forecasting and generation, we construct strong baselines based on state-of-the-art motion modeling practices.
LSTM-forecasting is a task-specific model that sequentially processes forecasting inputs $\{f_I(I_i), f_T(T_i)\}_{i=1:n}$ using an LSTM and outputs the motion $\bs{X}_{n+1:N}$.
Similarly, LSTM-generation processes $f_I(I_1)$ to generate $\bs{X}_{1:N}$.
We also train a two-stage model to replicate the two-stage approach used in prior works on wide-scene image-based forecasting~\cite{cao2020scenecontext} and generation~\cite{wang2021sceneaware}, but with diffusion modeling.
In particular, a UniEgoMotion-trajectory model first predicts the head trajectory, followed by the standard UniEgoMotion model, which takes the predicted head trajectory as additional input.
We further ablate our model by replacing the transformer decoder with a transformer encoder and a specialized 1D-UNet-based motion model~\cite{karunratanakul2023gmd}.
We also evaluate our motion representation against simple global representation~\cite{li2023egoego} and traditional pelvis-centric representation~\cite{guo2022generating,yi2024egoallo,jiang2022avatarposer}.
Finally, we examine the impact of using fine-grained features of DINOv2 versus text-optimized semantic features trained on general natural images (CLIP~\cite{radford2021clip}) and in-domain egocentric images (EgoVideo~\cite{pei2024egovideo}).

\subsection{Results \& Discussion}

\textbf{Egocentric Motion Reconstruction}:
Tab.~\ref{tab:recon_compare} shows the results and ablation study for the egocentric reconstruction task, and
Fig.~\ref{fig:recon} shows the qualitative comparison of the reconstruction baselines.
UniEgoMotion's egocentric reconstruction capabilities outperform specialized baselines -- AvatarPoser\cite{jiang2022avatarposer}, EgoEgo~\cite{li2023egoego}, and EgoAllo~\cite{yi2024egoallo}, in both reconstruction and semantic metrics.
EgoAllo's motion representation with a kinematic chain and local joint angles, often struggles to reconstruct motion accurately grounded on the floor.
As a result, it shows frequent foot sliding and floating motion, leading to its high Foot Contact error in Tab.~\ref{tab:recon_compare}.
We further verify the benefits of egocentric motion representation by training UniEgoMotion with a pelvis-centric representation.
We also show that transformer cross-attention is better suited for our flexible conditioning setting~\cite{rombach2022stablediff} compared to encoder-based architectures~\cite{li2023egoego,guzov2024hmd2} and specialized 1D-Unet~\cite{karunratanakul2023gmd}.
Thanks to the explicit use of fine-grained image features, UniEgoMotion captures visible semantic cues more effectively and achieves the lowest MPJPE-* errors along with the highest motion quality and semantic similarity to the ground-truth motion.

Although UniEgoMotion with CLIP~\cite{radford2021clip} image encoder shows strong performance~\cite{guzov2024hmd2}, the fine-grained features of DINOv2~\cite{oquab2023dinov2} are more suitable for extracting task-relevant scene context, which is often not the central focus of the image.
Surprisingly, the in-domain contrastive video features of EgoVideo~\cite{pei2024egovideo} perform slightly worse than CLIP, suggesting that generalized scene context is more important than ego-action centric image features.

\noindent\textbf{Egocentric Motion Forecasting \& Generation}:
Tab.~\ref{tab:fore_gen_compare} shows the quantitative results for the forecasting and generation tasks.
Qualitative results are shown in Fig.~\ref{fig:fore} \&~\ref{fig:gen} respectively.
Since the LSTM-based forecasting/generation baselines deterministically output the average of plausible future motions, they perform well on comparison metrics such as MPJPE and Semantic Similarity. However, the generated average motion exhibits significant foot-sliding and floor-penetration, resulting in lower motion quality, particularly affecting the Motion FID score.
The extensive two-stage baseline performs slightly worse on some metrics due to a mismatch between the distributions of the generated and groundtruth trajectories. However, it achieves strong motion quality metrics, benefiting from the vanilla UniEgoMotion model used in the second stage.
In comparison, UniEgoMotion enables one-shot high-quality forecasting and generation within a unified model, delivering strong overall performance across all metrics.

We also evaluate various design aspects of UniEgoMotion in the forecasting and generation tasks, and the results are consistent with those observed in the reconstruction task.
Motion representation plays an important role in generating high-quality motion, as reflected in metrics such as Foot Slide. The choice of image encoder has a slightly smaller impact than in reconstruction, as high-level scene context may be sufficient for generating/forecasting scene-relevant motion. However, a fine-grained image encoder still provides meaningful benefits in all metrics and noticeably improves motion realism, as observed in FC and FS metrics.
Interestingly, 1D-UNet~\cite{karunratanakul2023gmd}, which focuses on local motion reasoning, performs noticeably worse in both generation and forecasting.
For additional ablation and qualitative comparisons, please refer to the suppl. material.

\section{Conclusion \& Future Work}

We present UniEgoMotion, a unified framework for egocentric motion reconstruction, forecasting, and generation. Unlike previous methods, it extracts scene context from egocentric images, enabling scene-aware motion synthesis without explicit 3D scene. By integrating fine-grained visual features, our approach improves motion accuracy and realism. We also introduce EE4D-Motion, a large-scale dataset from EgoExo4D, offering time-synchronized egocentric video and pseudo-ground-truth 3D motion data. UniEgoMotion outperforms state-of-the-art methods in egocentric motion reconstruction while enabling novel egocentric forecasting and generation capabilities.
Our experiments emphasize the need for scene-aware motion reasoning. We show that instead of specialized motion architectures, a well-structured simple model with a strong context encoder and egocentric motion representation achieves superior results. Looking ahead, we plan to explore egocentric scene-motion interactions and leverage multimodal annotations for applications like in-context motion generation from text prompts. We believe UniEgoMotion provides a strong benchmark to drive future research in egocentric motion analysis and generation.

\noindent\textbf{Acknowledgment}
This research was conducted at Stanford University with support from Panasonic Holdings Corporation and partial funding from NIH grant R01AG089169.

{
    \small
    \bibliographystyle{ieeenat_fullname}
    \bibliography{main}
}


\clearpage
\maketitlesupplementary
\appendix

\begin{figure*}[h]
\centering
\includegraphics[width=\textwidth]{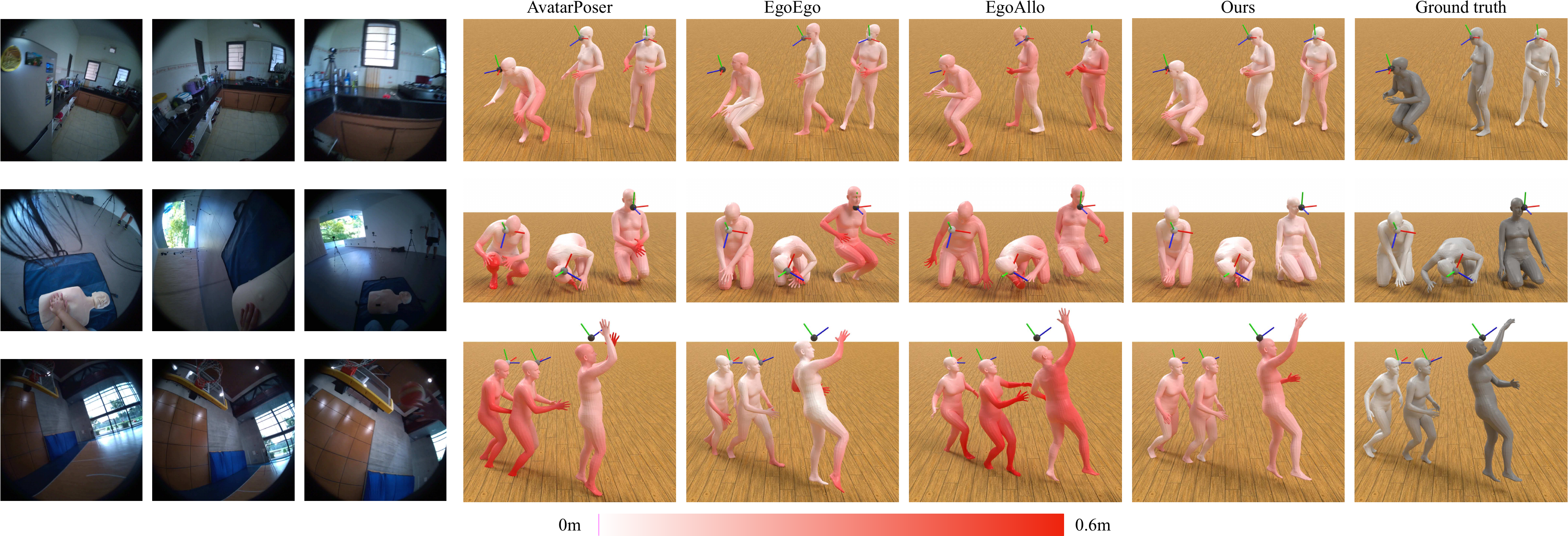}
\caption{Qualitative comparison of Egocentric Reconstruction, with absolute vertex errors color-coded. The input egocentric images are shown on the left, with the corresponding ego-device trajectory visualized alongside the predictions.}
\label{fig:error_recon}
\end{figure*}

\begin{table*}\centering \footnotesize
\setlength{\tabcolsep}{6.0pt}
\ra{1.2} 
\caption{\textbf{Ablation on Conditioning Inputs}: We evaluate UniEgoMotion in two ablation settings--without video and without trajectory input. Additionally, we train two single-modality variants of UniEgoMotion by conditioning only on trajectory or only on video.}
\begin{tabular}{lccccccccc}
\multicolumn{10}{c@{}}{Egocentric Motion Reconstruction} \\
\cmidrule{1-10}
\thead{Method} & \thead{Head Rot.\\ Err.} & \thead{Head Trans.\\Err.} &  \thead{MPJPE} & \thead{MPJPE-PA} & \thead{MPJPE-H} & \thead{Foot\\Slide} & \thead{Foot\\Contact} & \thead{Semantic\\Sim.($\uparrow$)} & \thead{FID} \\
\cmidrule{1-10}
UniEgoMotion & $\mathbf{0.260}$ &	$0.058$ &	$\mathbf{0.100}$ &	$\mathbf{0.053}$ &	$\mathbf{0.180}$ &	${3.62}$ &	$0.027$ &	$\mathbf{0.918}$ &	$0.027$ \\
\;\;\;\;w/o video & $0.278$ &	$\mathbf{0.057}$ &	$0.115$ &	$0.066$ &	$0.234$ &	$3.64$ &	$0.026$ &	$0.878$ &	$0.030$ \\
\;\;\;\;w/o trajectory & $0.539$ &	$0.280$ &	$0.290$ &	$0.059$ &	$0.352$ &	${2.95}$ &	$0.024$ &	$0.885$ &	$0.033$ \\
UniEgoMotion (w/o video) & $0.293$ &	$0.063$ &	$0.119$ &	$0.067$ &	$0.239$ &	$3.49$ &	$0.025$ &	$0.877$ &	$\mathbf{0.026}$ \\
UniEgoMotion (w/o trajectory) & $0.535$ &	$0.292$ &	$0.299$ &	$0.060$ &	$0.362$ &	$\mathbf{2.70}$ &	$\mathbf{0.023}$ &	$0.886$ &	$0.035$ \\
\bottomrule
\end{tabular}
\bigskip

\setlength{\tabcolsep}{3.2pt}
\begin{tabular}{l ccccccc|ccccccc}
\multicolumn{1}{c@{}}{} & \multicolumn{7}{c@{}}{Egocentric Motion Forecasting} & \multicolumn{7}{c@{}}{Egocentric Motion Generation} \\
\cmidrule{2-15}
\thead{Method} & \thead{J\\(2-4s)} & \thead{J-PA\\(2-4s)} & \thead{J-H\\(2-4s)} & \thead{FS} & \thead{FC} & \thead{SS ($\uparrow$)} & \thead{FID} &
\thead{J\\(0-2s)} & \thead{J-PA\\(0-2s)} & \thead{J-H\\(0-2s)} & \thead{FS} & \thead{FC} & \thead{SS ($\uparrow$)} & \thead{FID} \\
\cmidrule{1-15}
UniEgoMotion & $\mathbf{0.206}$ &	$0.071$ &	$\mathbf{0.308}$ &	${2.60}$ &	${0.026}$ &	$\mathbf{0.849}$ &	$\mathbf{0.047}$ 
& $\mathbf{0.226}$ &	$\mathbf{0.070}$ &	$\mathbf{0.321}$ &	${2.89}$ &	${0.025}$ &	${0.817}$ &	$\mathbf{0.043}$  \\
\;\;\;\;w/o video & $0.255$ &	$0.090$ &	$0.378$ &	$\mathbf{2.43}$ &	$0.028$ &	$0.782$ &	$0.058$
& $0.356$ &	$0.100$ &	$0.449$ &	$\mathbf{2.36}$ &	$0.027$ &	$0.696$ &	$0.065$  \\
\;\;\;\;w/o trajectory & $0.322$ &	$\mathbf{0.070}$ &	$0.414$ &	$2.66$ &	$0.025$ &	$0.838$ &	$\mathbf{0.047}$ 
& $\mathbf{0.226}$ &	$\mathbf{0.070}$ &	$\mathbf{0.321}$ &	$2.89$ &	$0.025$ &	$0.816$ &	$\mathbf{0.043}$  \\
UniEgoMotion (w/o video) & $0.276$ &	$0.095$ &	$0.400$ &	$2.69$ &	$0.028$ &	$0.767$ &	$0.067$
& $0.379$ &	$0.108$ &	$0.483$ &	$3.03$ &	$0.027$ &	$0.684$ &	$0.044$ \\
UniEgoMotion (w/o trajectory) & $0.318$ &	$\mathbf{0.070}$ &	$0.404$ &	$2.50$ &	$\mathbf{0.024}$ &	$0.842$ &	$0.050$
& $0.228$ &	$\mathbf{0.070}$ &	$\mathbf{0.321}$ &	$2.71$ &	$\mathbf{0.024}$ &	$\mathbf{0.820}$ &	$0.044$ \\
\bottomrule
\end{tabular}

\label{tab:cond_abl}

\end{table*}

\section{Qualitative Comparison}
See Fig.~\ref{fig:error_recon} for a qualitative visualization of egocentric motion reconstruction, with vertex errors color-coded. Please refer to the supplementary video to view UniEgoMotion's results on egocentric motion reconstruction, forecasting, and generation, as well as comparisons with baselines.

\section{Baselines}

\subsubsection*{Egocentric Motion Reconstruction}
We compare the egocentric motion reconstruction capabilities of UniEgoMotion with task-specific prior works: EgoEgo~\cite{li2023egoego}, EgoAllo~\cite{yi2024egoallo}, and AvatarPoser~\cite{jiang2022avatarposer}. To ensure a fair evaluation, we retrain each method on the EE4D-Motion dataset using their publicly available code. Following EgoEgo's experimental setup, we exclude hand tracking from AvatarPoser and instead provide a constant input for hand trajectories.
Both EgoEgo and AvatarPoser use as inputs the head trajectories derived from motion annotations rather than from the Aria device’s SLAM system, resulting in perfect head tracking by design. Similarly, EgoAllo also uses a fixed transformation between head and device trajectory. Therefore, we omit their head tracking metrics from the evaluation.
For EgoAllo, we evaluate the output of the motion diffusion model directly, without applying the post-processing optimization step.

Although EgoEgo and EgoAllo also adopt diffusion-based formulation for motion reconstruction, their approach differ from ours in their choice of motion representation and model architecture.
For instance, EgoEgo assumes a constant body shape and uses a global motion representation, whereas EgoAllo uses a head-centric representation that explicitly includes the head-to-pelvis transformation and preserves the kinematic chain.
More importantly, none of these baselines utilize semantic information from egocentric video for motion prediction. We compare these methods on the reconstruction task and also ablate their design choices separately within our UniEgoMotion framework in a consistent manner.

\subsubsection*{Egocentric Motion Forecasting \& Generation}
For egocentric motion forecasting and generation, the most relevant baselines~\cite{cao2020scenecontext,wang2021sceneaware} are two-stage models that generate or forecast human motion from third-person RGB images. They first predict the root trajectory (typically pelvis) and then generate the full-body human motion using a global motion representation.
To replicate these baselines faithfully, we train a separate UniEgoMotion variant that uses global motion representation and predicts only the root trajectory. This output is then provided as an additional conditioning input to the standard UniEgoMotion model (also with global motion representation) for full-body motion prediction.
We also train separate autoregressive LSTM-based baselines with a comparable model capacity for both forecasting and generation tasks. Since these models lack a generative component, their outputs tend to regress toward the mean of all plausible futures. As a result, they show lower error in direct comparison metrics such as MPJPE. However, their `averaged' prediction suffer from reduced motion diversity and realism, as shown in semantic metrics and qualitative visualization (see supplementary video).

\section{Ablation on Conditioning Inputs}
We evaluate UniEgoMotion under two ablation settings: without trajectory input and without video input. Additionally, we train two single-modality variants of UniEgoMotion. Egocentric reconstruction results in Tab.~\ref{tab:cond_abl} shows that both signals are useful for optimal reconstruction performance, thereby validating our use of video input, unlike prior baselines.
Interestingly, the separately trained single-modality variants offer no significant advantage over the original UniEgoMotion model when evaluated under the same conditions. 
Without video input, UniEgoMotion still outperforms baselines on most metrics. However, when the trajectory input is removed, the model is forced to implicitly solve visual odometry problem (a significantly harder task), leading to large errors on absolute metrics (head tracking, MPJPE, MPJPE-H). Despite this, it maintains accuracy in local pose metrics (MPJPE-PA, semantic similarity) and realism (FID), showing its ability to infer plausible motion from video alone.

EgoEgo~\cite{li2023egoego} employed an off-the-shelf monocular visual SLAM on egocentric video and trained an additional module to predict scale and the gravity vector to derive gravity-aligned metric SLAM trajectory. Their results showed that using predicted metric SLAM trajectory leads to only a minor degradation in pose metrics compared to using ground-truth trajectories.
In our work, we assume access to inertial SLAM trajectories for both our method and the baselines to decouple motion analysis from trajectory estimation and focus our evaluation on motion tasks. 

Forecasting and generation results follow similar trends, with both input modalities contributing to optimal performance.
Notably, the model without video input performs worse, as it lacks scene context necessary for generating or forecasting relevant motion.

\section{Motion Representation}

\begin{figure}[t]
\centering
\includegraphics[width=0.47\textwidth]{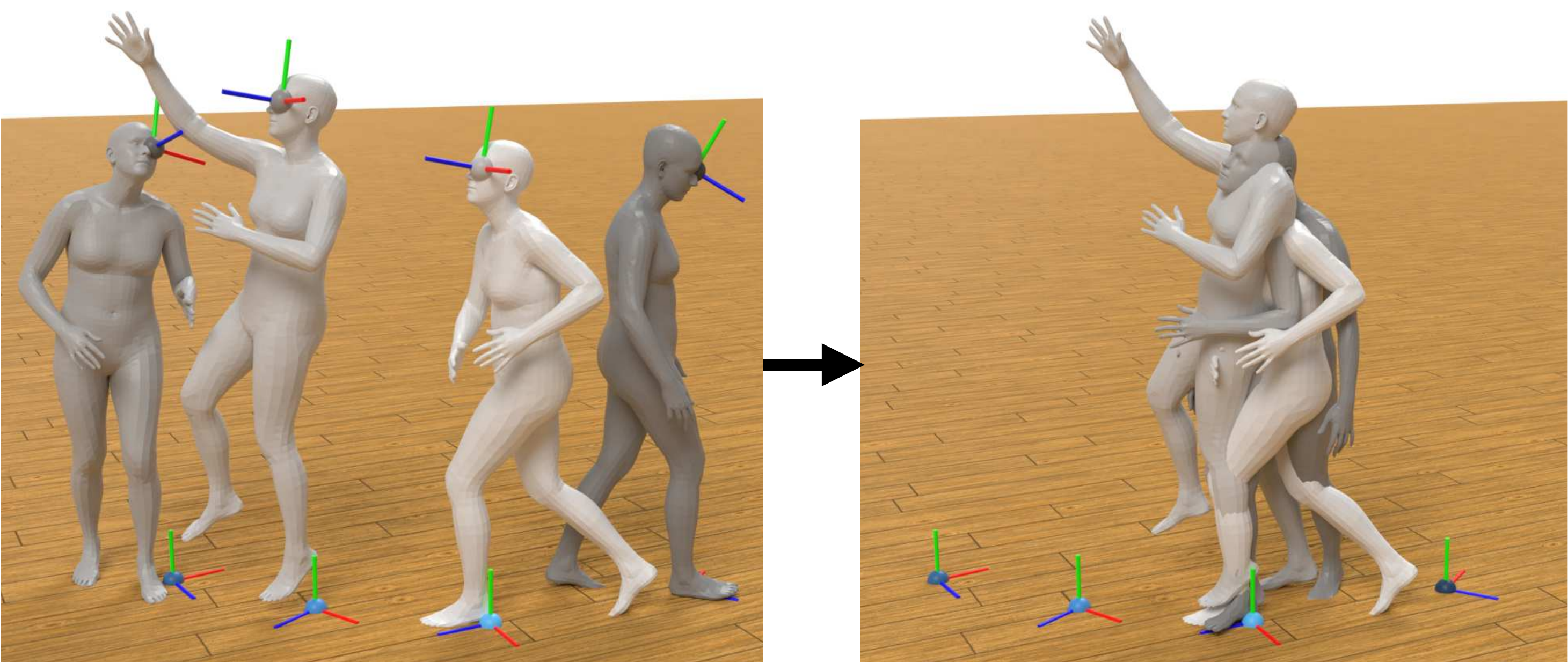}
\caption{Our egocentric motion representation decomposes motion into two components: (1) the egocentric trajectory projected onto the floor by removing pitch, roll, and height, and (2) the body pose relative to this projected egocentric trajectory.}
\label{fig:motion_repre}
\end{figure}

Although SMPL-X parameters $X_i = (R^r_i, t^r_i,  \theta_i, \beta_i)$ are sufficient to represent 3D body motion, they are not always ideal for learning~\cite{guo2022generating,yi2024egoallo}.
The global parameterization of the root trajectory $(R^r_i, t^r_i)$, defined at the pelvis, does not exploit motion invariances, forcing the model to learn all movements in every direction separately.
Moreover, a mismatch exists between the conditioning information $(T_i, I_i)$, defined in the egocentric frame, and the SMPL-X parameters $X_i$, defined in the pelvis-centric frame.
This misalignment complicates the reasoning between pelvis-centric motion and egocentric conditioning inputs.
Additionally, using local joint angles forces the model to reason complex forward kinematics of the SMPL-X skeleton, often resulting in suboptimal motion with noticeable artifacts such as foot-floor penetration and foot sliding.

To address these issues, we adopt a head-centric motion representation instead of a pelvis-centric one. We transform the SMPL-X parameters $X_i = (R^r_i, t^r_i,  \theta_i, \beta_i)$ into $(M^h_i, \bs{M}^j_i)$ using forward kinematics where $M^h_i \in \mb{R}^{4\times 4}$ is the global SE(3) transform of the head joint, and $\bs{M}^j_i \in \mb{R}^{21 \times 4\times 4}$ are the global SE(3) transforms of other joints.
This eliminates the dependency of each joint on its parent in the kinematic chain.
Next, we derive a canonical reference frame ${}_cM_i$ for each frame by projecting the head transform $M^h_i$ onto the floor.
In particular, ${}_cM_i$ represents the \textit{global} 3D transform of the head joint after removing the pitch and roll angle (keeping only yaw) and removing its height $t_z$ relative to the floor (+Z direction).
We then express the motion  $(M^h_i, \bs{M}^j_i)$ as $({}_cM_i, {}_cM_i \odot M^h_i, {}_cM_i \odot \bs{M}^j_i)$, where ${}_cM_i$ captures the head’s global trajectory projected onto the floor, and $({}_cM_i \odot M^h_i, {}_cM_i \odot \bs{M}^j_i)$ encode local canonicalized pose information.
To achieve trajectory invariance, we represent ${}_cM_i$ as its residual relative to the previous frame ${}_cM^{-1}_{i-1} \odot {}_cM_i$.
Following standard practice, we incorporate additional redundant information, such as joint locations and foot contact labels, into our motion representation.

While our motion representation is similar to the canonicalization in~\cite{yi2024egoallo}, it differs in that~\cite{yi2024egoallo} retains the kinematic chain and defines local joint rotations relative to parent joints. Since all body joint information in our approach is defined relative to the floor, it naturally facilitates better reasoning about foot-floor contact.
We validate the effectiveness of our motion representation through ablation studies and demonstrate that while ~\cite{yi2024egoallo} exhibits significant foot-floor penetration or floating artifacts, UniEgoMotion produces high-quality motion.

\section{Why Not Text Conditioning}
Many motion generation approaches~\cite{tevet2022mdm,EgoLM,wang2025ego4o} rely on text-based conditioning, where a clear textual prompt defines the intended motion or action. This explicit guidance simplifies the generation process. In contrast, our work focuses on passive conditioning using sensor data (e.g., video and device trajectory), where motion must be inferred without direct user input. While this introduces greater ambiguity, it also enables broader applicability in real-world scenarios such as continuous gait monitoring or fall prediction, where explicit user inputs are typically unavailable. Nonetheless, we believe that egocentric motion generation and forecasting from text prompts are promising future directions for many assistive applications. Our work, along with datasets like EE4D-Motion (with action narrations from EgoExo4D) and Nymeria~\cite{ma2024nymeria}, offers a promising starting point for such research.

\section{Training Details}
We train UniEgoMotion for 350 epochs using a batch size of 64 and the AdamW optimizer with a weight decay of 0.01. The learning rate is initialized at 3e-5 and decayed to 3e-6 after epoch 300. The model follows a standard transformer architecture~\cite{vaswani2017attention}, comprising 12 decoder layers with a latent dimension of 768. Training is conducted on 8-second motion sequences (80 steps at 10 fps), enabling long-horizon motion prediction. To improve training efficiency, DINOv2 features are precomputed and cached. End-to-end training takes approximately 2 days on a single NVIDIA L40S GPU. For diffusion, we use cosine noise scheduling with 1000 steps, consistent with prior works~\cite{tevet2022mdm,li2023egoego}, though effective motion synthesis has been demonstrated with very few diffusion steps~\cite{guzov2024hmd2,tevet2022mdm}. During training, we alternate between reconstruction and generation tasks with equal 0.5 probability by randomly masking the input sequence.

\section{EE4D-Motion Dataset}

Training UniEgoMotion requires paired egocentric videos and 3D human motion data within real-world environments. However, capturing 3D human motion in everyday activity settings—such as kitchens, offices, and sports fields—is challenging due to the cumbersome setup of motion capture systems. Existing large-scale 3D motion datasets~\cite{mahmood2019amass, lin2023motionx} lack paired egocentric videos, while most egocentric datasets either lack 3D motion annotations~\cite{grauman2022ego4d, grauman2024egoexo4d}, are small-scale~\cite{yuan2019egopose}, or have limited scene-motion correlation and diversity~\cite{li2023egoego,zhang2022egobody}.
The Nymeria dataset~\cite{ma2024nymeria} stands out with 200+ hours of daily activity egocentric videos paired with motion capture of simple skeleton sequences, but it does not provide the standard SMPL motion representation.

To bridge this gap, we process the large-scale EgoExo4D dataset~\cite{grauman2024egoexo4d} to generate pseudo-ground-truth 3D motion data. We refer to this processed dataset as EE4D-Motion, which consists of 208 hours of time-synchronized 3D motion data and egocentric videos, alongside other EgoExo4D annotations. This dataset serves as an extensive benchmark for multimodal motion research.

\subsection*{EgoExo4D Source Data}

EgoExo4D provides synchronized egocentric and exocentric video recordings of diverse activities, including cooking, dance, sports, music, healthcare, and bike repair. Egocentric videos were captured using Project Aria glasses~\cite{engel2023projectaria} along with the 3D trajectory of the ego camera. While EgoExo4D includes 3D body joint annotations for a subset of the dataset, these annotations are sparse, noisy, discontinuous, and lack joint angle information, making them unsuitable for motion tasks. Thus, we develop a processing pipeline to fit the SMPL-X body model to the continuous frames of EgoExo4D captures.

\subsection*{Fitting Pipeline}
Our pipeline leverages off-the-shelf models for pose estimation and follows a two-stage fitting approach~\cite{bogo2016smplify,lin2023motionx} to obtain 3D-accurate motion groundtruth. We exclude rock climbing sequences to focus on motions occurring on a flat surface. Our pipeline consists of the following steps.

\noindent\textbf{Detection \& Tracking}: We detect~\cite{li2022vitdet} and track the egocentric camera wearer in each exo view. When multiple people are present, we use the Aria 3D trajectory to identify the person of interest.

\noindent\textbf{Pose Estimation}: For each bounding box, we estimate 2D keypoints~\cite{xu2022vitpose} and obtain an initial SMPL-X parameter estimate using an off-the-shelf HMR model~\cite{cai2023smplerx}. However, single-view HMR estimates suffer from depth ambiguity and jitter in 3D translation.

\noindent\textbf{Per-Frame Fitting}: We initialize SMPL-X fitting by averaging HMR estimates across exo views. The fitting optimizes SMPL-X parameters $(R^r, t^r, \theta, \beta)$ using the following energy term~\cite{bogo2016smplify}:
\begin{align*}
\mc{L}_\text{fitting} &= \lambda_\theta E_\theta(\theta) + \lambda_\beta E_\beta(\beta) \\
&+ \lambda_{2d} \sum_v \sigma\left(\pi_v(J(R^r, t^r, \theta, \beta)) - K^{2d}_v\right)
\end{align*}
where $E_\theta$ and $E_\beta$ are priors for pose and shape, respectively, $J$ is the SMPL-X 3D joint regressor, $\pi_v$ is the 2D projection operator using known camera intrinsics and extrinsics of view $v$, $K^{2d}_v$ represents detected 2D joints, $\sigma$ is the robust Geman-McClure function~\cite{gmloss, bogo2016smplify}, and $\lambda_*$ are energy weights.

\noindent\textbf{Sequence-Level Optimization}: After per-frame fitting, we refine results at the sequence level by fixing the body shape $\beta$ as the average across the sequence, incorporating egocentric view detections, and adding a temporal jitter penalty to enforce smooth motion.

\noindent\textbf{Filtering \& Quality Control}: We filter out segments with excessive jitter caused by erroneous device trajectories, suboptimal off-the-shelf model predictions, or severe occlusions across all exo views. After filtering, we retain 110 hours of smooth and accurate EE4D-Motion data for UniEgoMotion training.

Through this pipeline, EE4D-Motion provides 3D-accurate motion annotations aligned with egocentric video, enabling us to train and evaluate UniEgoMotion model.

\subsection*{Motion Annotations Quality}
EE4D-Motion annotations can be noisy in scenes with poor exocentric visibility (e.g., kitchen, COVID testing) or large camera distances (e.g., basketball). EgoExo4D’s own pose annotations are sparse and jittery, resulting in high pose error of $\sim$0.24m for EgoEgo, as reported by the authors of EgoExo4D~\cite{grauman2024egoexo4d}, compared to  $\sim$0.16m on our smoother and denser annotations. Unlike EgoEgo’s synthetic dataset, where motions are scene-agnostic, EE4D-Motion provides contextually grounded motion aligned with real-world environments, which is essential for both generation and forecasting tasks.



\end{document}